\documentclass{article}
\usepackage{spconf,amsmath,graphicx,amssymb,amsthm,amsmath,mathtools}

\usepackage{subfigure,epstopdf,algorithm}
\usepackage{array}
\usepackage{xcolor}
\usepackage{thmtools}

\newtheorem{corollary}{Corollary}

\declaretheorem{proposition}

\usepackage{times}
\usepackage{eqparbox,algorithm,algpseudocode}

\def\x{{\mathbf x}}

\def\t{{\mathbf t}}

\def\T{{\mathbf T}}
\def\F{{\mathbf G}}
\def\F{{\mathbf F}}
\def\Q{{\mathbf Q}}
\def\v{\mbox{vec}}
\title{Recursive prediction of graph signals with incoming nodes}
\name{Arun~Venkitaraman$^*$, Saikat Chatterjee$^\dagger$, Bo Wahlberg$^*$\thanks{This work was supported by the Wallenberg AI, Autonomous Systems and Software Program (WASP) and the Swedish Research Council (2016-06079)}}

\address{
	$^*$Division of Decision and Control Systems, KTH Royal Institute of Technology, Stockholm, Sweden        \\
	$^\dagger$Division of Information Science and Engineering, KTH Royal Institute of Technology, Stockholm, Sweden   \\
	\{arunv,sach,bo\}@kth.se
}
\begin{document}
	\ninept
	\maketitle
	\begin{abstract}
		Kernel and linear regression have been recently explored in the prediction of graph signals as the output, given arbitrary input signals that are agnostic to the graph. In many real-world problems, the graph expands over time as new nodes get introduced. Keeping this premise in mind, we propose a method to recursively obtain the optimal prediction or regression coefficients for the recently proposed Linear Regression over Graphs (LRG), as the graph expands with incoming nodes. This comes as a natural consequence of the structure of the regression problem, and obviates the need to solve a new regression problem each time a new node is added. Experiments with real-world graph signals show that our approach results in good prediction performance which tends to be close to that obtained from knowing the entire graph apriori.
	\end{abstract}
	\begin{keywords}
		Linear regression, graph expansion, graph signal processing, recursive least squares, convex optimization.
	\end{keywords}
	
	\section{Introduction}
	Graph signal processing (GSP) has recently emerged as a general framework for characterization and analysis of data over graphs or networks \cite{gsp_overview_ortega}. A large part of the existing GSP literature may be classified into two groups: first, where the graph is completely known apriori and is employed in signal analysis, and the second, where the graph or topology is estimated from the signals. In both the cases, either the number of nodes or the edges of the entire graph is assumed fixed apriori. This is not the case in many practical systems where the nodes may be introduced thereby expanding an existing graph. It is then natural to ask if the existing knowledge is transferrable in an {\it online} manner when the graph expands: for a given task, it is desirable to use the solution obtained for the current graph as a bias or initialization for the new graph obtained by adding a node to the current graph. In this work, we address this problem for the recently proposed technique of Linear Regression over Graphs (LRG) \cite{Arun_krg_tsipn}. 
	
	Given an input $\mathbf{x}\in\mathbb{R}^I$ associated to a target $\mathbf{y}\in\mathbb{R}^M$, LRG constructs the following model:
	\begin{equation}
	\mathbf{y}=\mathbf{W}^\top\pmb\phi(\mathbf{x}),\,\,\mathbf{W}\in\mathbb{R}^{K\times M} \nonumber
	\vspace{-.2in}
	\end{equation}
	where
	\begin{itemize}
		\item the input $\mathbf{x}$ is agnostic to a graph (is not assumed to be a graph signal); $\pmb\phi$ is a known feature map which operates on the input $\mathbf{x}$ to generate the feature vector $\pmb\phi(\mathbf{x})\in\mathbb{R}^K$,
		\item the predicted target or output $\mathbf{y}$ is a smooth graph signal over a given graph $\mathcal{G}$ of $M$ nodes,
	%	\item $\mathcal{G}=(\mathcal{V},\mathcal{E},\mathbf{A})$ where $\mathcal{V}$ denotes the node set, $\mathcal{E}$ the edge set, and $\mathbf{A}=[a_{ij}]$ the adjacency matrix, $a_{ij}\geq0$
		%\item a training data-set of $N$ input-output pairs $\{(\mathbf{x}_n,\mathbf{t}_n)\}_{n=1}^N$, and
		\item the training targets $\mathbf{t}_n$ are measurements of smooth graph signals $\mathbf{y}_{n,o}$ through additive white noise $\mathbf{e}_n$ uncorrelated with the signal:
		%	\begin{eqnarray}
		$\quad\mathbf{t}_n=\mathbf{y}_{n,o}+\mathbf{e}_n$.
		%\nonumber
		%\end{eqnarray}
		%\item $\mathbf{W}\in\mathbb{R}^{K\times M} $
	\end{itemize}
Given a training dataset of $N$ input-output pairs $\{(\mathbf{x}_n,\mathbf{t}_n)\}_{n=1}^N$, the regression coefficient matrix $\mathbf{W}$ is computed by minimizing the following cost function:
	\begin{align}
\label{eq:elm_cost_2}
C(\mathbf{W})=\sum_{n=1}^N \|\mathbf{t}_n-\mathbf{y}_n\|_2^2 + \alpha \|\mathbf{W}\|_F^2+\beta \sum_{n=1}^N \mathbf{y}_n^\top\mathbf{L}\mathbf{y}_n\nonumber\\
\mbox{subject to  }\mathbf{y}_n=\mathbf{W}^\top\pmb\phi(\x_n)\,\,\forall\,n,\quad\,\,\alpha,\beta\geq0,
\end{align}
where $\|\|_F$ denotes the Frobenius norm and $\mathbf{L}$ is the graph-Laplacian associated with $\mathcal{G}$. The details of this cost and LRG are explained in Section 1.3. The optimal regression coefficients $\hat{\mathbf{W}}$ are then a function of the graph-Laplacian matrix $\mathbf{L}$, the targets and input features \cite{Arun_krg_tsipn}. LRG has been shown to be the state-of-the-art in the prediction of graph signals with significant gains when the training data is corrupted with noise or is scarce. Further, it is the only method which is capable of handling the case where the input is not assumed to be a graph signal. 

Given this premise, we ask if LRG can be extended to the case where the graph expands as new nodes get introduced. More specifically, we ask if the optimal regression coefficients $\hat{\mathbf{W}}_{M}$ obtained for a given graph $\mathcal{G}_M$ of $M$ nodes could be used to derive $\hat{\mathbf{W}}_{M+1}$ for the graph $\mathcal{G}_{M+1}$ obtained by introducing a node to $\mathcal{G}_M$. Towards this end, our main contributions in this work are as follows:
\begin{itemize}
	\item We derive a recursive algorithm to update the optimal LRG coefficients $-$ {\it obviates} the need to solve a new batch problem every time each time a new node is introduced
	\item We apply our framework, which we term {\it node-recursive LRG (NR-LRG)}, to real-world datasets to validate our theory. 
	%$-$ that the recursive framework can be used to arrive at the coefficients for the expanded graph.
\end{itemize}

\subsection{Literature survey}
	Graph signal processing offers a consistent treatment of graph-structured data in a wide variety of applications, be it in social networks, traffic networks, or biological networks \cite{Shuman,Sandry2} As mentioned in the beginning of this Section, GSP literature may be classified into two groups: one of them being that which uses an apiori specified graph for signal processing tasks over graphs. This includes a wealth of techniques from harmonic and filterbank analysis \cite{windowedGFT,Narang2013,ArunGHT,Coifman2006}, sampling \cite{graphsamp10,graphsamp11,graphsamp6}, statistical analysis \cite{statgraph2_journal}, non-parametric analysis \cite{Arun_krg_tsipn,kergraph1,multikernel_Arun}, prediction and recovery \cite{chen1,kergraph3}  to the more recent graph neural networks \cite{gcnn_surver}. The second group deals with the problem of graph estimation or discovery where the graph signals are used to arrive at an estimate of the graph or the connections among the nodes \cite{dong2018learning,glkrg,Arun_graphlearn_SPL}. 
	%
%	In many real-world scenarios, one often encounters graphs or networks which slowly expand over time with the entry of new nodes. For example, the case of a social network where a new individual joins an existing circle of friends. In such cases, most of the existing GSP techniques would have to be re-calculated or re-solved every time the graph expands with even one additional node. It is natural to ask if the existing knowledge is transferrable in an {\it online} manner when the graph expands: for a given task, it is desirable to use the solution obtained for the current graph as a bias or initialization for the new graph obtained by adding a node to the current graph. In this work, we solve this problem for the task of linear regression over graphs (LRG) which deals with prediction of smooth graph signals as output for a given input that is not necessarily on the same graph \cite{Arun_krg_tsipn}. We derive a recursive method of obtaining the regression coefficients (which we term {\em node-recursive linear regression over graphs (NR-LRG)}) that are updated each time a new node is added to the graph. The exact manner in which the weights are updated depends on the edge from the incoming node. 
Recursive approaches have been considered in the context of distributed recursive least squares by Mateo et al. \cite{rls_distributed,rls_consensus}. Diffusion recursive least squares for distributed estimation was investigated by Cattivelli  et al. \cite{rls_adaptive_networks}. {Di Lorenzo} et al. \cite{rls_adaptive_gsp} recently considered recursive least squares in the adaptive reconstruction of graph signals where the updates take place over time and across the nodes of a fixed graph/network topology. 
	
	We next briefly review the basics of graph signal processing, followed by a review of linear regression over graphs.
	\subsection{Graph signal processing}
	Consider a graph of $M$ nodes denoted by $\mathcal{G}=(\mathcal{V},\mathcal{E},\mathbf{A})$ where $\mathcal{V}$ denotes the node set, $\mathcal{E}$ the edge set, and $\mathbf{A}=[a_{ij}]$ the adjacency matrix, $a_{ij}\geq0$. Since we consider only undirected graphs, we have that $\mathbf{A}$ is symmetric \cite{Chung}. A vector $\mathbf{y}=[y(1) y(2) \cdots y(M)]^\top\in\mathbb{R}^{M}$ is said to be a graph signal over $\mathcal{G}$ if $y(m)$ denotes the value of the signal at the $m$th node of the graph \cite{Shuman,Sandry2,gsp_overview_ortega}. The smoothness of a graph signal $\mathbf{y}$ is measured in terms of the quadratic form:
	\begin{equation*}\mathbf{y}^\top\mathbf{L}\mathbf{y}=\sum_{(i,j)\in\mathcal{E}}a_{ij}\left[y(i)-y(j)\right]^2,
	\end{equation*}
	where
	$\mathbf{L}=\mathbf{D}-\mathbf{A}$ is the graph Laplacian matrix, 
	$\mathbf{D}$ being the diagonal degree matrix with $i$th diagonal given by $d_{i}=\sum_ja_{ij}$. 
	$\mathbf{y}^\top\mathbf{L}\mathbf{y}$ is a measure of variation of $\mathbf{y}$ across connected nodes: the smaller the value, the smoother the signal $\mathbf{y}$. 
	
	\subsection{Linear regression for graph signals (LRG)}
	As discussed earlier, the optimal LRG coefficients are obtained by minimizing  $C(\mathbf{W})$ in \eqref{eq:elm_cost_2}.
	The first term in $C(\mathbf{W})$ denotes the training error. The second term ensures that the regression coefficients remain bounded specially at low sample sizes. Lastly, the third regularization enforces the regression output to be smooth over the graph. 
	Using the properties of the matrix trace operation $\mbox{tr}(\cdot)$ and on applying further simplifications on \eqref{eq:elm_cost_2}, we get that \cite{Arun_krg_tsipn}
	\begin{align}
	\label{eq:elm_g_cost_function_w}
	C(\mathbf{W}) 
	& = \mbox{tr}(\T^\top\T)-2\,\mbox{tr}\left(\mathbf{T}^\top\pmb\Phi\mathbf{W} \right) +\mbox{tr}\left(  \mathbf{W}^\top\pmb\Phi^\top\pmb\Phi\mathbf{W}\right)\nonumber \\
	& \quad+ \alpha\, \mbox{tr}(\mathbf{W}^\top\mathbf{W}) +  \beta\, \mbox{tr}\left(  \mathbf{W}^\top\pmb\Phi^\top\pmb\Phi\mathbf{W}\mathbf{ L}\right),
	% \nonumber\\
	\end{align}
	where $\pmb\Phi=[\pmb\phi(\mathbf{x}_1)\cdots\pmb\phi(\mathbf{x}_N)]^\top\in\mathbb{R}^{N\times K}$ and $\mathbf{T}=[\mathbf{t}_1\cdots\mathbf{t}_N]^\top$. Note that as long as the input $\mathbf{x}$ remains the same, $\pmb\Phi$ does  not depend on the graph or $M$. The optimal LRG coefficient matrix $\hat{\mathbf{W}}$, obtained by setting $\frac{\partial C}{\partial\mathbf{W}}=\mathbf{0}$, is given by \cite{Arun_krg_tsipn}:
	\begin{align}
	\label{eq:nrlrg_w}
	\mbox{vec}({\hat{\mathbf{W}}}) \! = \!  \mathbf{F}^{-1}   (\mathbf{I}\otimes\pmb\Phi^\top)\mbox{vec}(\mathbf{T})
	\end{align}
	where $\mbox{vec}(\cdot)$ denotes the vectorization operation, $\otimes$ the Kronecker product, and the matrix $\mathbf{F}=[\mathbf{I}_{M}+ \beta \mathbf{L}]\otimes \pmb\Phi^\top\pmb\Phi+\alpha\mathbf{I}_{KM}$. We note that the solution for the regression coefficients obtained thus assumes knowledge of the entire graph apriori. In the next Section, we show how to arrive at a node-recursive formulation for LRG.
	
	\section{ Node-Recursive Linear Regression for Graphs} 
	Let us now consider the case of two undirected graphs $\mathcal{G}_M=(\mathbf{A}_M,\mathcal{V}_M,\mathcal{E}_M)$ and  $\mathcal{G}_{M+1}=(\mathbf{A}_{M+1},\mathcal{V}_{M+1},\mathcal{E}_{M+1})$ with $M$ and $M+1$ nodes, respectively. The graph $\mathcal{G}_{M+1}$ is obtained by inserting one additional node $\mathcal{G}_M$ without affecting the existing edges. In terms of the adjacency matrices, we have that
	\begin{equation}
	\mathbf{A}_{M+1}=\left[
	\begin{array}{cc}
	\mathbf{A}_M&\mathbf{a}_{M+1}\\
	\mathbf{a}_{M+1}^\top& 0
	\end{array}\right]\nonumber
	\end{equation}
	where $\mathbf{a}_{M+1}=[a_{M+1,1},a_{M+1,2},\cdots,a_{M+1,M}]^\top$, and we have assumed no self-loops at the nodes, that is, $a_{m,m}=0\,\forall m$. Let $\mathbf{L}_M$ and $\mathbf{L}_{M+1}$ denote the corresponding graph-Laplacian matrices.
	Further, let $\hat{\mathbf{W}}_M$ denote the optimal regression coefficient matrix obtained with $N$ training samples for $\mathcal{G}_M$, and $\hat{\mathbf{W}}_{M+1}$ for $\mathcal{G}_{M+1}$. In order to differentiate between the target vectors of the two graphs, we use $\mathbf{T}_n^M$ and $\mathbf{T}_n^{M+1}$ for graphs $\mathcal{G}_M$ and $\mathcal{G}_{M+1}$, respectively. Then, $(\mathbf{x}_n,\mathbf{t}_{n}^M)$ denotes the $n$th input-output pair for $\mathcal{G}_M$, and $(\mathbf{x}_n,\mathbf{t}_{n}^{M+1})$ denotes the $n$th input-output pair for $\mathcal{G}_{M+1}$, respectively. Then, we have that \begin{equation*}
	\mathbf{t}_n^{M+1}=[\t_n^M\, t_n(M+1)]^\top,
	\end{equation*} where $t_n(M+1)$ denotes the target value at the new node corresponding to index $M+1$ for the $n$th training sample. Let $\mathbf{\underbar{t}}^{M+1}=[t_1(M+1), t_2(M+1),\cdots t_N(M+1)]^\top\in\mathbb{R}^N$ denote the vector of $N$ signal values at the $(M+1)$th (incoming) node for all observations from $1$ to $N$. 
	\begin{proposition} The optimal regression coefficient matrix for  $\mathcal{G}_{M+1}$ follows the recursive relation:
		\begin{align}
		&\mbox{vec}({\hat{\mathbf{W}}_{M+1}})=\left[\begin{array}{c}
		(\mathbf{I}_{MK}-\pmb\rho_M)\,\v(\hat{\mathbf{W}}_M)+\mathbf{m}_{M+1}\pmb\Phi^\top\mathbf{\underbar{t}}^{M+1}\\
		\mathbf{m}^\top_{M+1}\F_M\,\v(\hat{\mathbf{W}}_M)+\mathbf{n}_{M+1}\pmb\Phi^\top\mathbf{\underbar{t}}^{M+1}
		\end{array}\right].\nonumber
		\end{align}
		where $(\mathbf{I}_{MK}-\pmb\rho_M)$, $\mathbf{m}_{M+1}$, $\mathbf{F}$, and $\mathbf{n}_{M+1}$ are weight matrices that depend only upon $\mathbf{a}_{M+1}$, and  $\pmb\Phi$. 
	\end{proposition}
	\noindent We present the details and the proof of Proposition 1 in Section 5, and describe the NR-LRG algorithm in Algorithm 1. 
	
	We make two important observations from the Proposition: First, the updated regression coefficients in $\hat{\mathbf{W}}_{M+1}$ are a weighted combination of the values $\hat{\mathbf{W}}_{M}$ and the new node data $\mathbf{\underbar{t}}^{M+1}$. Intuitively, we expect that the change in regression coefficients corresponding to the first $M$ nodes would be proportional to the influence of the incoming node: stronger the links $\mathbf{a}_{M+1}$, more the influence of on the regression matrix. Secondly, we expect that if the incoming node has weak edges, regression coefficients corresponding to $(M+1)$th node should give higher importance to its own data. In the case of a completely disconnected node, one should expect that no change occurs in the weights of the prior $M$ nodes, and the regression $(M+1)$th node should be independent of $\mathcal{G}_M$ (it should correspond to the solution of the regularized least squares or the ridge-regression). This is indeed the case, and is a direct corrollary of Proposition 1:
	\begin{corollary}
		In the case when the new node is not connected with those in $\mathcal{G}$, that is $\mathbf{a}_{M+1}=\mathbf{0}$, we have that
		\begin{align}
		&\mbox{vec}({\hat{\mathbf{W}}_{M+1}})=\left[\begin{array}{c}
		\v(\hat{\mathbf{W}}_M)\\
		(\pmb\Phi^\top\pmb\Phi+\alpha\mathbf{I}_K)^{-1}\pmb\Phi^\top\mathbf{\underbar{t}}^{M+1}
		\end{array}\right].\nonumber
		\end{align}
	\end{corollary}
	\noindent This shows that the algorithm makes no changes to the regression coefficients of the existing graph if the incoming node has no edges. The regression coefficients for the new node then simply correspond to those obtained from the individual regularized least squares or ridge regression.
	\subsection{ The choice of the hyperparameters $\alpha$ and $\beta$}
	We here note that the NR-LRG recursions are derived assuming that the hyperparameters $\alpha,\,\beta$ do not change as the graph expands. In practice however, the hyperparameters are set using cross-validation. As a result, the best $\alpha$ and $\beta$ will usually vary from graph of one size to the other. Hence, in actual practice, the recursive solution would not always be equal to the batch solution obtained by using the entire graph. This is evidenced from the results obtained on real-world datasets in Section 4 $-$ the performance curves of LRG and NR-LRG almost coincide but not exactly.
		\begin{algorithm}[t] % enter the algorithm environment
		\caption{Node-recursive Linear Regression over Graphs} % give the algorithm a caption
		\label{alg1} % and a label for \ref{} commands later in the document
		\begin{algorithmic}[1] % enter the algorithmic environment
			\State Initialize $M=M_0$, 
			\State Compute $\hat{\mathbf{W}}_M$ using \eqref{eq:nrlrg_w}, \State Set $\mathbf{Q}_M=(\mathbf{I}_M \! \otimes \! (\pmb\Phi^\top\pmb\Phi+\alpha\mathbf{I})+\beta \mathbf{L}_M \! \otimes \! \pmb\Phi^\top\pmb\Phi)^{-1}$
			\While{$M\leq M_{max}$}
			\State$\mathbf{h}_{M+1}=\beta\mathrm{diag}(\mathbf{a}_{M+1})\otimes \pmb\Phi^\top\pmb\Phi-\mathbf{c}_{M+1}\mathbf{d}_{M+1}^{-1}\mathbf{c}^\top_{M+1}$ \qquad
			\State$\pmb\rho_M=\mathbf{h}_{M+1}(\Q_{M}-\pmb\rho_M\Q_M)$
			\State $\Q_{M+1}=\left[\begin{array}{cc}
			\Q_{M}-\pmb\rho_M\Q_M&\mathbf{m}_{M+1}\\
			\mathbf{m}^\top_{M+1}&\mathbf{n}_{M+1}
			\end{array}\right]$\\
			$\mbox{vec}({\hat{\mathbf{W}}_{M+1}})=\left[\begin{array}{c}
			(\mathbf{I}_{MK}-\pmb\rho_M)\v(\hat{\mathbf{W}}_M)+\mathbf{m}_{M+1}\pmb\Phi^\top\mathbf{t}^{M+1}\\
			\mathbf{m}^\top_{M+1}\F_M\v(\hat{\mathbf{W}}_M)+\mathbf{n}_{M+1}\pmb\Phi^\top\mathbf{t}^{M+1}
			\end{array}\right]$
			\EndWhile
		\end{algorithmic}
	\end{algorithm}
	\section{Experiments}
	We consider the application on two real-world graph signal datasets. In these datasets, the true targets $\mathbf{y}_{o,n}$'s are smooth graph signals which lie over a specified graph. During the training phase, the targets are observed with additive white Gaussian noise at a particular signal-to-noise-ratio (SNR) level. We apply Algorithm 1 to obtain the train the LRG model, starting from an initial graph of size $M_0$. The trained models are then used to predict targets for inputs from the test dataset. We compare the test prediction performance using the normalized-mean square error:
	$NMSE=\displaystyle\frac{\mbox{E}\|\mathbf{y}_{n,o}-\mathbf{y}\|_2^2}{\mbox{E}\|\mathbf{y}_{n,o}\|_2^2}$. The expected value is obtained as the average over different datapoints and noise realizations.  Each time a node is added, we compute the NMSE on the test data of the corresponding $M+1$ nodes. We compare the performance of NR-LRG to that of linear regression (LR) (which does not use the graph structure), and linear regression over graphs (LRG). The hyperparameters for LR and LRG are obtained through four-fold cross-validation. We use the values of $\alpha$ and $\beta$ obtained from LRG when solving NR-LRG. 
	\subsection{Temperature prediction on Swedish cities}
\begin{figure*}[t]
		\vspace{-.1in}
		\centering
		$
		\begin{array}{cc}
		\subfigure[]{
			\includegraphics[width=2.2in]{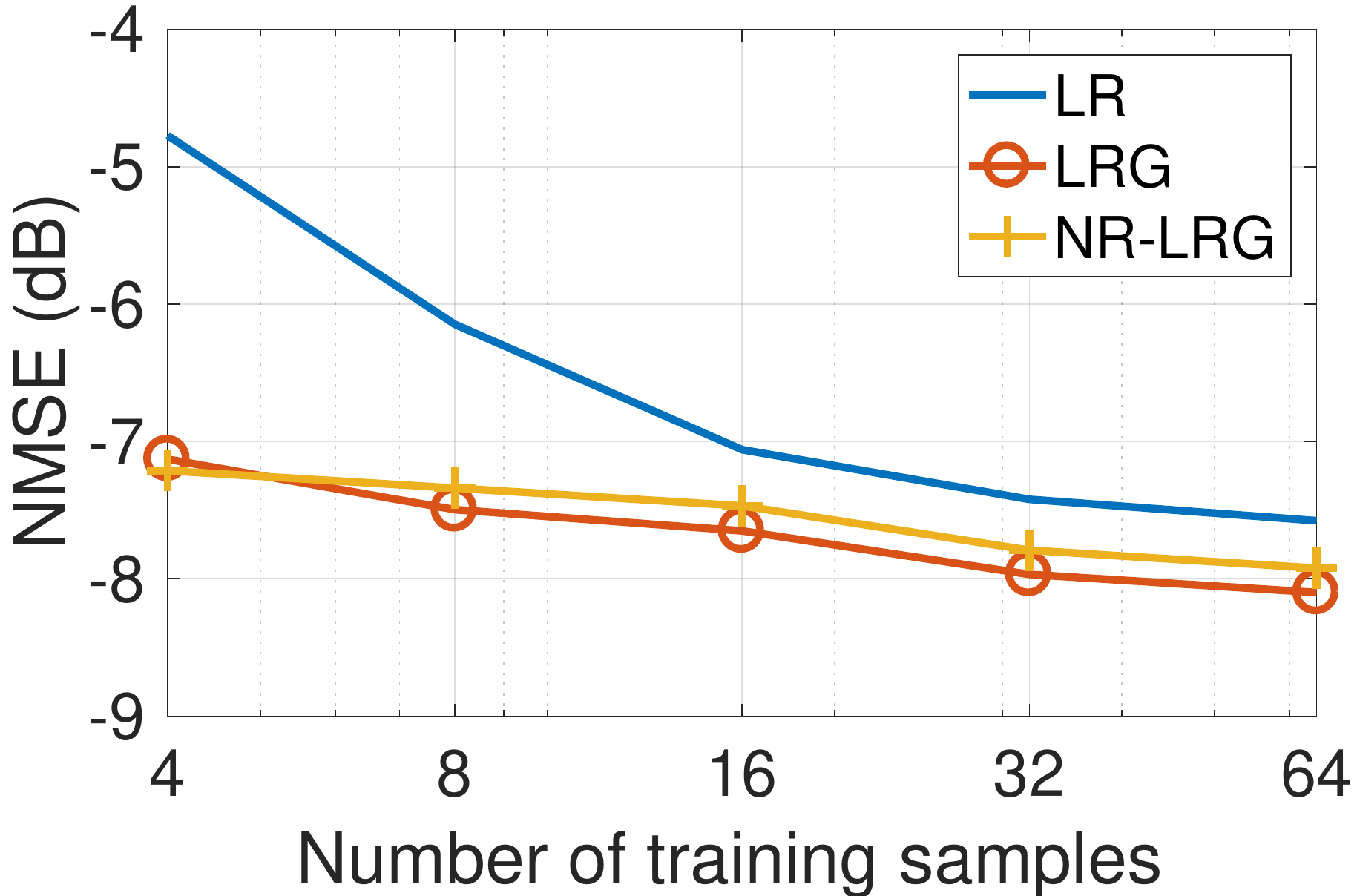}}
		\subfigure[]{
			\includegraphics[width=2.2in]{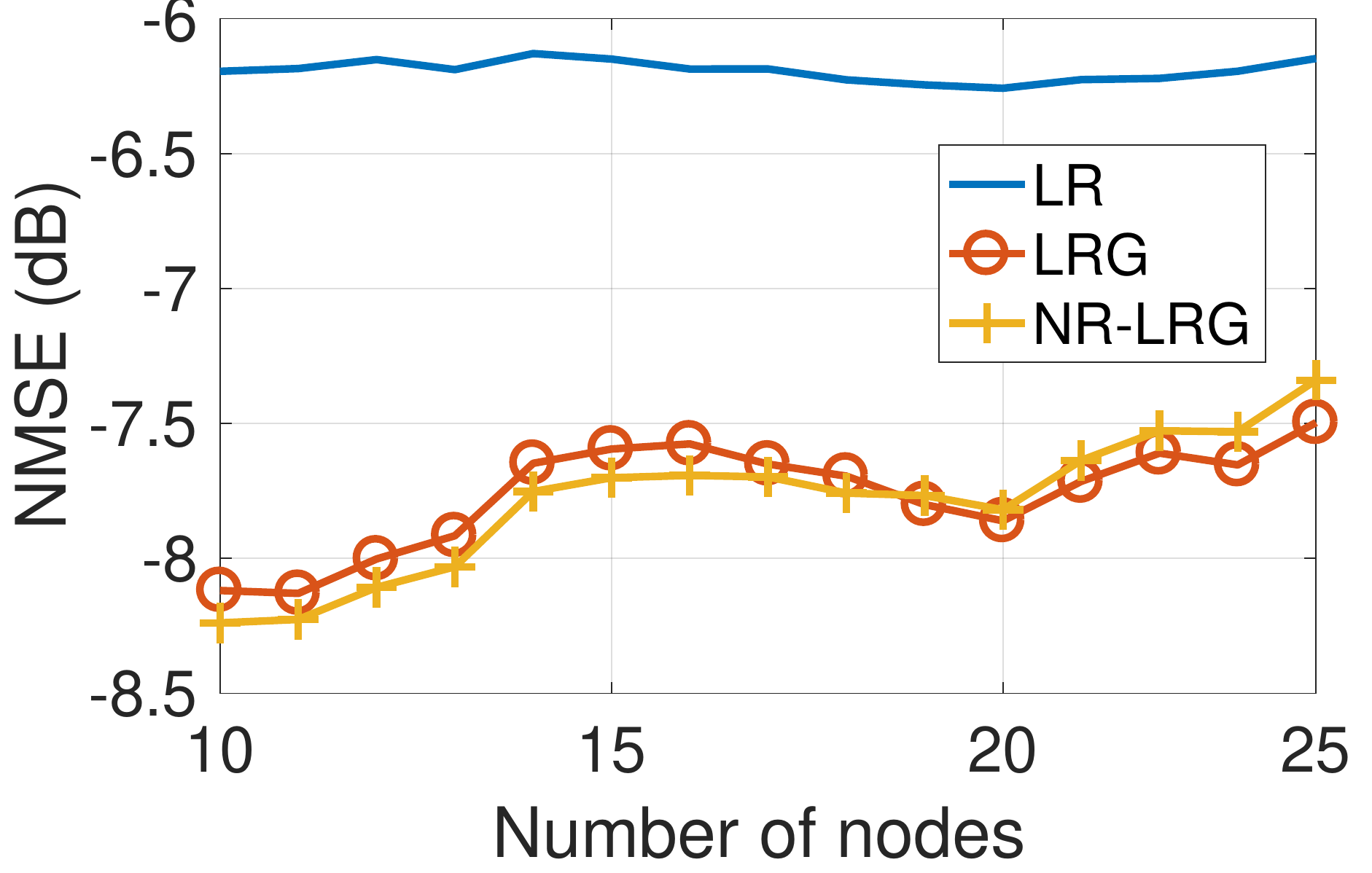}}
		\subfigure[]{\includegraphics[width=2.0in]{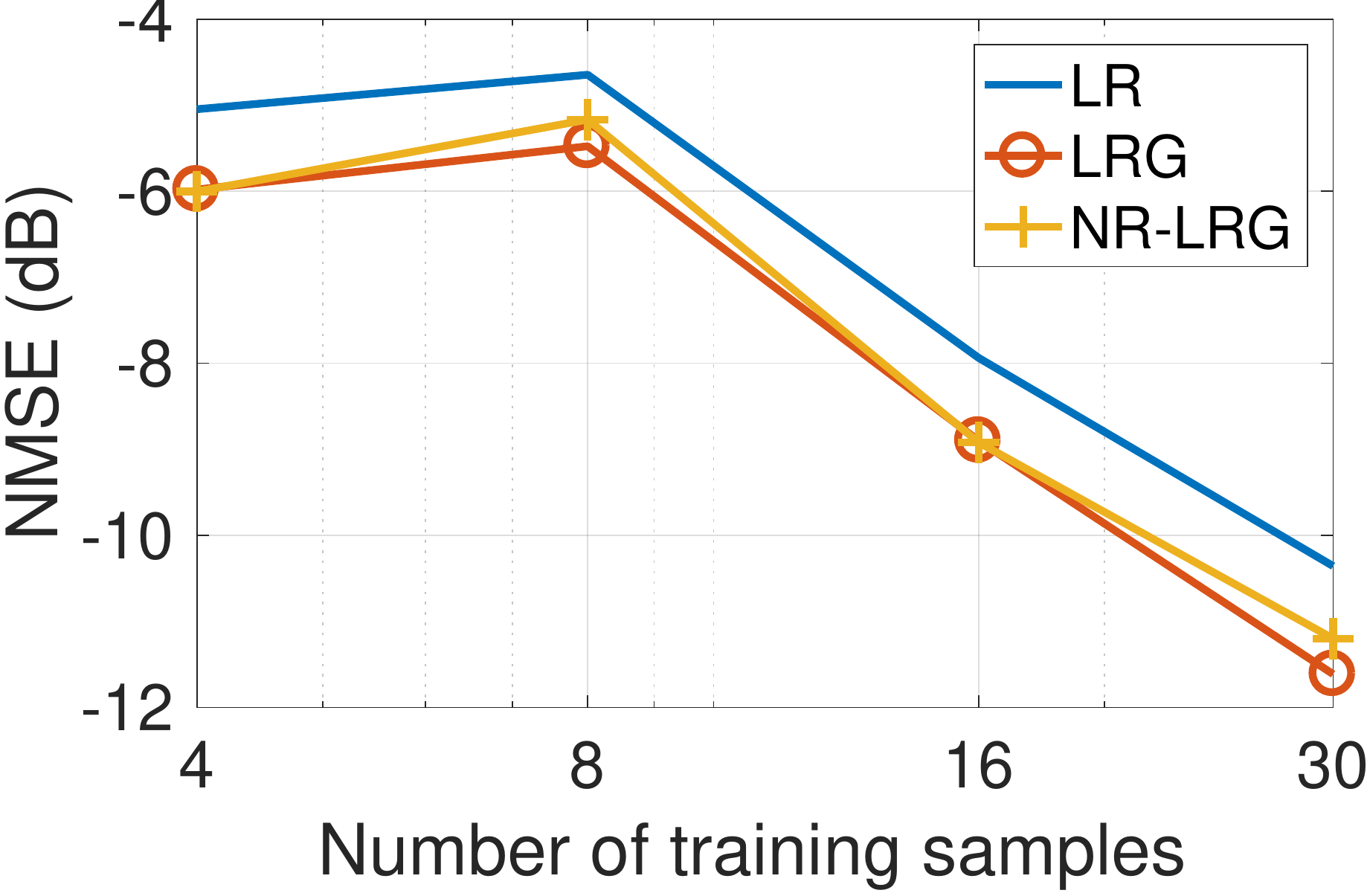}}
		\end{array}$
		\caption{NMSE performance on test data (a) for the temperature prediction experiment as a function of training sample size $N$, (b) for the temperature prediction experiment as a function of incoming nodes $M$, when $N=8$, and (c) for the ETEX dataset as a function of $N$.}
		\label{fig:nrlrg_temp}
	\end{figure*}

	We first consider the problem of two-day temperature prediction on $25$ of the most populated cities in Sweden. The data was collected over the period of October to December 2017 and was used in \cite{Arun_krg_tsipn}. The input $\mathbf{x}\in\mathbb{R}^{25}$ comprises the temperature values across the cities on a given day, and the predicted target $\mathbf{y}\in\mathbb{R}^{25}$ corresponds to the temperature values after two days. The total dataset thus contains $90$ input-output pairs. We use the first 64 as the training set and the remaining for testing. In this experiment, we consider the identity input feature map $\pmb\phi(\mathbf{x})=\mathbf{x}$. The entire graph of $M=25$ nodes corresponding to the cities is defined through the adjacency matrix $a_{i,j}=\exp{\left(-\frac{d_{ij}^2}{\sum_{i,j}d_{ij}^2}\right)}$, where $d_{ij}$ denotes the geodesic distance between the $i$th and $j$th cities. We consider additive white noise at an SNR level of $10$dB. We initialize NR-LRG with the subgraph of size $M_0=5$ and run the algorithm till $M=25$. This means that at each recursion the temperature readings from all the cities is known, but the output changes dimensions from $M=10$ to  $M=25$. In Figure \ref{fig:nrlrg_temp}(a), we show the NMSE obtained on the test data for LR, LRG and, NR-LRG at the different training data-sizes, when the NR-LRG is initialized with $M_0=10$. We observe that the performance of NR-LRG and LRG are close to each other and almost coincide when the number of training datapoints is small. The test prediction performance of NR-LRG as a function of nodes with the number training data-points set to $N=8$ is shown in Figure \ref{fig:nrlrg_temp}(b). We observe that NR-LRG predictions are close to that obtained from LRG using the data from the entire set of $M=25$ nodes, and clearly outperform LR where no graph information is used.
	
	%
%	\begin{figure}
%		\centering
%		\includegraphics[width=2.0in]{nmse_nrlrg_etex10.pdf}
%		\caption{NMSE on test data for the ETEX experiment.}
%		\label{fig:nrlrg_etex}
%	\end{figure}
	
	\subsection{Prediction of atmospheric pollution index}
	We next consider the application of NR-LRG to atmospheric particulant density data from the European Tracer Experiment (ETEX). The data consists of daily readings of the particulant density at various cities over Europe taken over period of two months or for 60 days. We take as the input the readings at a subset $S_{in}$ of 30 cities and the target predicted as the readings at a different set $S_{op}$ of 30 cities. As with the earlier experiment, we consider the graph defined with adjacency matrix based on the geodesic distances. Further, we consider a nonlinear input feature map given by $\pmb\phi(\mathbf{x})=[\phi_1(\mathbf{x})\cdots\phi_K(\mathbf{x})]$, where:
	\begin{equation*}
	\pmb\phi_i(\mathbf{x})=\displaystyle\frac{1}{1+\exp(-(\mathbf{f}_i^\top\x+{g}_i))}
	\end{equation*}
	and $\mathbf{f}_i$s and ${g}_i$s are randomly drawn from the normal distribution. This represents a random nonlinear transformation of the input which has been shown to be of merit in various neural network architectures \cite{elm_HUANG2010,elmg_2}. We have used $K=2M$, which means the input feature map is twice the dimension of the input. We initialize NR-LRG with $M_0=5$, and consider training data at an SNR-level of $10$dB. Thus, each at the $m$th recursion on NR-LRG, input is of dimension $30$ corresponding to $S_{in}$ and the output corresponds to the first $m$ cities in $S_{op}$. The prediction NMSE of LR, LRG, and NR-LRG on test data are shown in Figure \ref{fig:nrlrg_temp}(c). Once again, we observe that the NR-LRG performs similarly to the batch LRG.
	
	\section{Discussion and Conclusions}
	We proposed an approach to recursively obtain the regression coefficients for linear regression over graphs. The recursion makes it possible to obtain updated regression coefficients as a new node is introduced to an existing graph. Formulating a recursive approach makes it possible to update the regression 'online' without having the need to know the graph in its entirety apriori. This makes our approach particularly suited to realistic graph signal scenarios, where the graph often expands over time.
	
	We note however, that though our experiments on real-world datasets revealed empirically that the node-recursive version performs similarly to the non-recursive complete approach, it is not clear if such a claim can be made in a mathematically rigorous manner. One of the challenges in pursuing such an investigation is the growing dimension of the parameters being estimated. Standard recursive approaches assume that the parameter that is updated in an online manner is of fixed dimension and meaning: we do not have that advantage. In our future work, we hope to explore this aspect further.
	\vspace{-.1in}
	\section{Proof of Proposition 1}
%	We now proceed to prove the proposition. We split the proof into the following sub-sections to enhance the readability.
%	\subsection{The graph-Laplacian matrices}
	By definition, the degree matrix $\mathbf{D}=\mbox{diag}(\mathbf{A}\mathbf{1})$ is a diagonal matrix with the sum of rows $\mathbf{A}$. Then, we have that for $i\leq M$
	\begin{align}
	\mathbf{D}_{M+1}(i,i)=\sum_{j=1}^{M+1}a_{i,j}=\sum_{j=1}^{M}a_{i,j}+a_{i,M+1}=\mathbf{D}_{M}(i,i)+a_{i,M+1}.\nonumber
	\end{align}
	%Putting in matrix form, we have
	%\begin{equation}
	%\mathbf{D}_{M+1}=\left[
	%\begin{array}{cc}
	%\mathbf{D}_M+\mathrm{diag}(\mathbf{a}_{M+1})&\mathbf{0}\\
	%\mathbf{0}& \sum_{j=1}^Ma_{M+1,j}.\nonumber
	%\end{array}\right]
	%\end{equation}
	This in turn leads to the following relation between $\mathbf{L}_M$ and $\mathbf{L}_{M+1}$:
	\begin{equation}
	\mathbf{L}_{M+1}=\left[
	\begin{array}{cc}
	\mathbf{L}_M+\mathrm{diag}(\mathbf{a}_{M+1})&-\mathbf{a}_{M+1}\\
	-\mathbf{a}^\top_{M+1}& \mathbf{a}_{M+1}^\top\mathbf{1}_M
	\end{array}\right],\nonumber
	\end{equation}
	since $\sum_{j=1}^Ma_{M+1,j}=\mathbf{a}^\top_{M+1}\mathbf{1}_M$, where $\mathbf{1}$ denotes the vector with all ones.
	%\subsection{Regression coefficient matrix}
	From \eqref{eq:nrlrg_w}, we have that the output weight matrix for $M+1$ nodes is given by
	\begin{align}
	\mbox{vec}({\hat{\mathbf{W}}_M})=\F_M^{-1}  \mbox{vec}(\pmb\Phi^\top\mathbf{T}_M)
	\! = \!  \F_M^{-1}   (\mathbf{I}\otimes\pmb\Phi^\top)\mbox{vec}(\mathbf{T}_M)\nonumber
	\end{align}
	where $\F_M=[\mathbf{I}_{M}+ \beta \mathbf{L}_{M}]\otimes \pmb\Phi^\top\pmb\Phi+\alpha\mathbf{I}_{KM}$. Then, we have that
	\begin{align}
	\label{eq:nrlrg_proof_c_d}
	\F_{M+1}&=
	%\mathbf{I}_{M+1} \otimes  (\pmb\Phi^\top\pmb\Phi+  \alpha\mathbf{I}_K) + \beta \mathbf{L}_{M+1}\otimes  \pmb\Phi^\top\pmb\Phi\nonumber\\
	[\mathbf{I}_{M+1}+ \beta \mathbf{L}_{M+1}]\otimes \pmb\Phi^\top\pmb\Phi+\alpha\mathbf{I}_{K(M+1)}\nonumber\\
	%&=\left[\begin{array}{cc}
	%\mathbf{I}_M+\beta\mathbf{L}_M&-\beta\mathbf{a}_{M+1}\\
	%+\beta\mathrm{diag}(\mathbf{a}_{M+1})&\\
	%&\\
	%-\beta\mathbf{a}^\top_{M+1}& 1+\beta\mathbf{a}^\top_{M+1}\mathbf{1}_M
	%\end{array}\right]\otimes %\pmb\Phi^\top\pmb\Phi+\alpha\mathbf{I}_{K(M+1)}\nonumber\\
	&=\left[\begin{array}{cc}
	(\mathbf{I}_M+\beta\mathbf{L}_M)\otimes \pmb\Phi^\top\pmb\Phi&\\
	+\alpha\mathbf{I}_{MK}&-\beta\mathbf{a}_{M+1}\otimes \pmb\Phi^\top\pmb\Phi\\
	+\beta\mathrm{diag}(\mathbf{a}_{M+1})\otimes \pmb\Phi^\top\pmb\Phi&\\
	&\\
	-\beta\mathbf{a}^\top_{M+1}\otimes \pmb\Phi^\top\pmb\Phi& (1+\beta\mathbf{a}^\top_{M+1}\mathbf{1}_M)\pmb\Phi^\top\pmb\Phi\\
	&+\alpha\mathbf{I}_K
	\end{array}\right]\nonumber\\
	&=\left[\begin{array}{cc}
	\F_M&-\beta\mathbf{a}_{M+1}\otimes \pmb\Phi^\top\pmb\Phi\\
	+\beta\mathrm{diag}(\mathbf{a}_{M+1})\otimes \pmb\Phi^\top\pmb\Phi&\\
	&\\
	-\beta\mathbf{a}^\top_{M+1}\otimes \pmb\Phi^\top\pmb\Phi& (1+\beta\mathbf{a}^\top_{M+1}\mathbf{1}_M) \pmb\Phi^\top\pmb\Phi\\
	&+\alpha\mathbf{I}_K
	\end{array}\right]\nonumber\\
	&\triangleq\left[\begin{array}{cc}
	\mathbf{b}_{M+1} &\mathbf{c}_{M+1}\\
	\mathbf{c}^\top_{M+1}&\mathbf{d}_{M+1}
	\end{array}\right],
	\end{align}
	where we have used the commutativity property of the Kronecker product \cite{Loan1}.
	%where 
	%\begin{enumerate}
	%	\item[] $\mathbf{b}_{M+1}=\F_{M}+\beta\mathrm{diag}(\mathbf{a}_{M+1})\otimes \pmb\Phi^\top\pmb\Phi$
	%	\item[] $\mathbf{c}_{M+1}=-\beta\mathbf{a}_{M+1}\otimes \pmb\Phi^\top\pmb\Phi $
	%	\item[] $\mathbf{d}_{M+1}=(1+\beta\mathbf{a}^\top_{M+1}\mathbf{1}_M) \pmb\Phi^\top\pmb\Phi+\alpha\mathbf{I}_K$.
	%\end{enumerate}
	%
	%
	%
	%
	%
%	\subsection{Inverse of the regression matrix}
	Taking the inverse of $\mathbf{F}_{M+1}$ and applying Sherman-Morrison-Woodbury formulae, we have that\\
	%\begin{align}
	%\label{eq:nrlrg_proof_m_n}
	%
	$[\F_{M+1}]^{-1}=\left[\begin{array}{cc}
	\mathbf{z}_{M+1} &\mathbf{m}_{M+1}\\
	\mathbf{m}^\top_{M+1}&\mathbf{n}_{M+1}
	\end{array}\right],\nonumber
	$
	%\end{align}
	where
	\begin{enumerate}
		\item[]$\mathbf{z}_{M+1} =(\mathbf{b}_M-\mathbf{c}_{M+1}\mathbf{d}_{M+1}^{-1}\mathbf{c}^\top_{M+1})^{-1}$
		\item[]$\mathbf{m}_{M+1} =-\mathbf{z}_{M+1}\mathbf{c}_{M+1}\mathbf{d}_{M+1}^{-1}$
		\item[]$\mathbf{n}_{M+1} =\mathbf{d}_{M+1}^{-1}-\mathbf{d}_{M+1}^{-1}\mathbf{c}^\top_{M+1}\mathbf{z}_{M+1}\mathbf{c}_{M+1}\mathbf{d}_{M+1}^{-1}$
	\end{enumerate}
	Let $\mathbf{Q}_M\triangleq[\F_M]^{-1}$. Then,
	the matrix $\mathbf{z}_{M+1}$ is simplified as
	\begin{align}
	\mathbf{z}_{M+1}
	%&=(\mathbf{b}_M-\mathbf{c}_{M+1}\mathbf{d}_{M+1}^{-1}\mathbf{c}^\top_{M+1})^{-1}\nonumber\\
	&=\left[\F_{M}+\beta\mathrm{diag}(\mathbf{a}_{M+1})\otimes \pmb\Phi^\top\pmb\Phi-\mathbf{c}_{M+1}\mathbf{d}_{M+1}^{-1}\mathbf{c}^\top_{M+1}\right]^{-1}\nonumber\\
	&{=}\left[\Q_{M}^{-1}+\mathbf{h}_{M+1}\right]^{-1}
	%&=\left[\F_{M}+\beta\mathrm{diag}(\mathbf{a}_{M+1})\otimes \pmb\Phi^\top\pmb\Phi-\beta^2[\mathbf{a}_{M+1}\otimes \pmb\Phi^\top\pmb\Phi] \mathbf{d}_{M+1}^{-1}[\mathbf{a}^\top_{M+1}\otimes \pmb\Phi^\top\pmb\Phi]\right]^{-1}\nonumber\\
	%&=(\mathbf{I}+\beta\mathbf{L}_M)^{-1}-(\mathbf{I}+\beta\mathbf{L}_M)^{-1}[(\mathbf{h}^{-1}_{M+1}+(\mathbf{I}+\beta\mathbf{L}_M)^{-1}]^{-1}(\mathbf{I}+\beta\mathbf{L}_M)^{-1}\nonumber\\
	%&=[\F_{M}]^{-1}-[\F_M]^{-1}(\mathbf{h}^{-1}_{M+1}+[\F_M]^{-1})^{-1}[\F_M]^{-1}\nonumber\\
	%&=\Q_{M}-\Q_M(\mathbf{h}^{-1}_{M+1}+\Q_M)^{-1}\Q_M
	=\Q_{M}-\pmb\rho_M\Q_M,\nonumber
	\end{align}
	where $\mathbf{h}_{M+1}=\beta\mathrm{diag}(\mathbf{a}_{M+1})\otimes \pmb\Phi^\top\pmb\Phi-\mathbf{c}_{M+1}\mathbf{d}_{M+1}^{-1}\mathbf{c}^\top_{M+1}$ and the matrix $\pmb\rho_M$ is given by
	\begin{align}
	&	\pmb\rho_M=\Q_M(\mathbf{h}^{-1}_{M+1}+\Q_M)^{-1},\quad\mbox{or}\quad	\pmb\rho_M(\mathbf{h}^{-1}_{M+1}+\Q_M)=\Q_M\nonumber\\
	&	\pmb\rho_M\mathbf{h}^{-1}_{M+1}=\Q_M-\pmb\rho_M\Q_M,\quad\mbox{or}\quad	\pmb\rho_M=\mathbf{h}_{M+1}\mathbf{z}_{M+1}.
	\end{align}
	Similarly, expressing the matrices $\mathbf{m}_{M+1}$ and $\mathbf{n}_{M+1}$ in terms of $\Q_M$:
	\begin{align}
	\mathbf{m}_{M+1}
	%&=-(\mathbf{b}_M-\mathbf{c}_{M+1}\mathbf{d}_{M+1}^{-1}\mathbf{c}^\top_{M+1})^{-1}\mathbf{c}_{M+1}\mathbf{d}_{M+1}^{-1}\nonumber\\
	&=-[\Q_{M}-\pmb\rho_M\Q_M]\mathbf{c}_{M+1}\mathbf{d}_{M+1}^{-1},\,\,\,\mbox{and}\\
	\mathbf{n}_{M+1}
	%&=\mathbf{d}_{M+1}^{-1}-\mathbf{d}_{M+1}^{-1}\mathbf{c}^\top_{M+1}\mathbf{z}_{M+1}\mathbf{c}_{M+1}\mathbf{d}_{M+1}^{-1}\nonumber\\
	&=\mathbf{d}_{M+1}^{-1}-\mathbf{d}_{M+1}^{-1}\mathbf{c}^\top_{M+1}[\Q_{M}-\pmb\rho_M\Q_M]\mathbf{c}_{M+1} \mathbf{d}_{M+1}^{-1}\nonumber
	\end{align}
	Then, we get that the inverse of $\F_{M+1}$is then given by
	\begin{align}
	\label{eq:elmg_L_recurs_g_2}
	\F^{-1}_{M+1}\triangleq\Q_{M+1}=\left[\begin{array}{cc}
	\Q_{M}-\pmb\rho_M\Q_M &\mathbf{m}_{M+1}\\
	\mathbf{m}^\top_{M+1}&\mathbf{n}_{M+1}
	\end{array}\right]
	\end{align}
	Also, we have that $\v(\mathbf{T}_{M+1})=\left[\begin{array}{rr}\v(\mathbf{T}_{M+1})\\
	\mathbf{\underbar{t}}^{M+1}\end{array}\right]$, where $\mathbf{\underbar{t}}^{M+1}$ denotes the vector of $N$ observations at the $(M+1)$th node.
	\subsection{Deriving the final recursion on the coefficients}
	Putting the relevant equations together, we have that
	\begin{align}
	&\mbox{vec}({\mathbf{W}_{M+1}})=
	\F_{M+1}^{-1}   (\mathbf{I}_{(M+1)K}\otimes\pmb\Phi^\top)\left[\begin{array}{rr}\v(\mathbf{T}_{M})\\
	\mathbf{\underbar{t}}^{M+1}\end{array}\right]\nonumber\\
	&= \F_{M+1}^{-1}   \left[\begin{array}{cc}
	\mathbf{I}_{MK}\otimes\pmb\Phi^\top&0\\
	0& \pmb\Phi^\top\end{array}\right]\left[\begin{array}{rr}\v(\mathbf{T}_{M})\\
	\mathbf{\underbar{t}}^{M+1}\end{array}\right]\nonumber\\
	&=\F_{M+1}^{-1}   \left[\begin{array}{rr}(\mathbf{I}_{MK}\otimes\pmb\Phi^\top)\v(\mathbf{T}_{M})\\
	\pmb\Phi^\top\mathbf{\underbar{t}}^{M+1}\end{array}\right]=\F_{M+1}^{-1}   \left[\begin{array}{rr}\F_M\v(\mathbf{W}_M)\\
	\pmb\Phi^\top\mathbf{\underbar{t}}^{M+1}\end{array}\right]\nonumber\\
	&=\left[\begin{array}{cc}
	\Q_{M}-\pmb\rho_M\Q_M &\mathbf{m}_{M+1}\\
	\mathbf{m}^\top_{M+1}&\mathbf{n}_{M+1}
	\end{array}\right]\left[\begin{array}{rr}\F_M\v(\mathbf{W}_M)\\
	\pmb\Phi^\top\mathbf{\underbar{t}}^{M+1}\end{array}\right]\nonumber\\
	&=\left[\begin{array}{c}
	(\mathbf{I}_{MK}-\pmb\rho_M)\Q_{M}\F_M\v(\mathbf{W}_M)+\mathbf{m}_{M+1}\pmb\Phi^\top\mathbf{\underbar{t}}^{M+1}\\
	\mathbf{m}^\top_{M+1}\F_M\v(\mathbf{W}_M)+\mathbf{n}_{M+1}\pmb\Phi^\top\mathbf{\underbar{t}}^{M+1}
	\end{array}\right]\nonumber\\
	&=\left[\begin{array}{c}
	(\mathbf{I}_{MK}-\pmb\rho_M)\v(\mathbf{W}_M)+\mathbf{m}_{M+1}\pmb\Phi^\top\mathbf{\underbar{t}}^{M+1}\\
	\mathbf{m}^\top_{M+1}\F_M\v(\mathbf{W}_M)+\mathbf{n}_{M+1}\pmb\Phi^\top\mathbf{\underbar{t}}^{M+1}
	\end{array}\right]
	%	&\triangleq\left[\begin{array}{c}
	%	\mathbf{R}_{1,M+1}\v(\mathbf{W}_M)+\mathbf{R}_{2,M+1}\pmb\Phi^\top\mathbf{\underbar{t}}^{M+1}\\
	%	\mathbf{R}_{3,M+1}\v(\mathbf{W}_M)+\mathbf{n}_{M+1}\pmb\Phi^\top\mathbf{\underbar{t}}^{M+1}
	%	\end{array}\right].\nonumber
	\end{align}
	%where 
	%
	%\begin{enumerate}
	%\item[] $\mathbf{D}_{1,M+1}=	(\mathbf{I}_{MK}-\pmb\rho_M)$
	%\item[] $\mathbf{D}_{2,M+1}=\beta[\Q_{M}-\pmb\rho_M\Q_M][\mathbf{a}_{M+1}\otimes \pmb\Phi^\top\pmb\Phi]\mathbf{d}_{M+1}^{-1}$
	%\item[]
	%	$\mathbf{D}_{3,M+1}=\beta\mathbf{d}_{M+1}^{-1}[\mathbf{a}_{M+1}\otimes \pmb\Phi^\top\pmb\Phi] [\mathbf{I}_{MK}-\pmb\rho_M]$
	%\end{enumerate}
%
	In the case when the incoming node is disconnected, $\mathbf{a}_{M+1}=\mathbf{0}$. It can then be verified that $\pmb\rho_M=0$, $\mathbf{m}_{M+1}=\mathbf{0}$ and $\mathbf{n}_{M+1}=(\pmb\Phi^\top\pmb\Phi+\alpha\mathbf{I}_K)^{-1}$, thereby
	proving Corollary 1.

	\clearpage
	
	\bibliographystyle{IEEEbib}
\bibliography{refs,FinalBibArun,bibliography}
	
\end{document}